\documentclass{article}
\usepackage{spconf,amsmath,graphicx}
\usepackage{multirow}

\title{Facial Expressions as a Vulnerability in Face Recognition}
\twoauthors{A. Pe\~na, A. Morales, I. Serna, J. Fierrez}
{Biometrics and Data Pattern Analytics Lab\\Universidad Autonoma de Madrid, Spain}
{A. Lapedriza}{Computer Science Dpt. / eHealth Center \\ Universitat Oberta de Catalunya, Spain\\}
\begin{document}
%
\maketitle
\begin{abstract}
This work explores facial expression bias as a security vulnerability of face recognition systems. Despite the great performance achieved by state-of-the-art face recognition systems, the algorithms are still sensitive to a large range of covariates. We present a comprehensive analysis of how facial expression bias impacts the performance of face recognition technologies. Our study analyzes: i)  facial expression biases in the most popular face recognition databases; and ii) the impact of facial expression in face recognition performances. Our experimental framework includes two face detectors, three face recognition models, and three different databases. Our results demonstrate a huge facial expression bias in the most widely used databases, as well as a related impact of face expression in the performance of state-of-the-art algorithms. This work opens the door to new research lines focused on mitigating the observed vulnerability.
\end{abstract}
\section{Introduction}\label{sec:introduction}

The progress of face recognition techniques have allowed practical uses of these technologies, such as access control in private places (e.g. gyms or office buildings), safely unlocking your smartphone, or systems that facilitate automatic identity verification in the airports’ border control. With the emergence of deep learning architectures, the accuracy of these technologies have largely surpassed human-level performance~\cite{patel18spm_face}. However, despite these impressive results, some studies show that deep learning based face recognition systems remain significantly sensitive to different sources of variation in face images, commonly referred as covariates~\cite{2018_TIFS_SoftWildAnno_Sosa,lu2019experimental}. This has motivated researchers to study ways to reduce the impact of these factors for years. 

\begin{figure*}[t]
\centering
\includegraphics[width=0.95\textwidth]{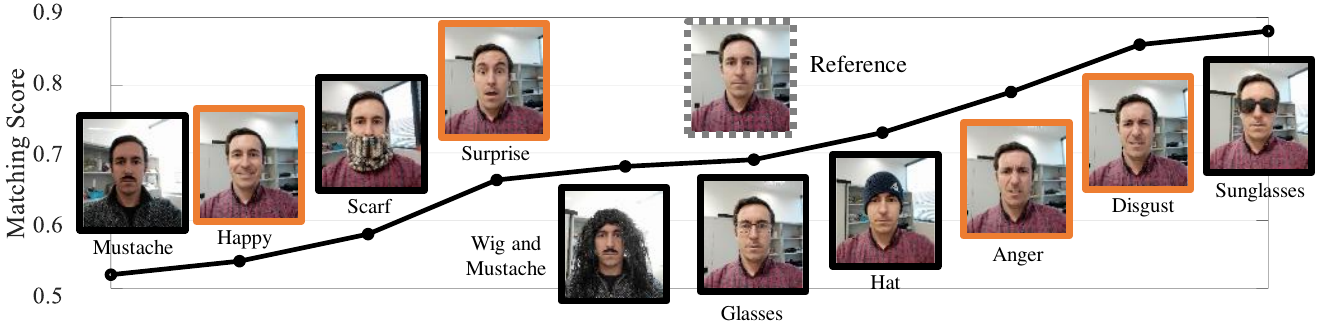} 
\caption{Matching score (Euclidean distance computed on ResNet-$50$~\cite{ResNet} embeddings) between a Reference image (grey dotted line) and 10 Test images including different facial expressions (orange line) and accesories to obfuscate the face (black line).}
\label{fig:changes_app}
\end{figure*}

On a different front, researchers have demonstrated that face recognition technologies are vulnerable to different types of attempts to intervene the automatic process of recognition, which have been traditionally studied as \emph{attacks}~\cite{Galbally2007_Vulnerabilities}. These attacks can be divided into digital or physical attacks. Digital attacks are those intended to manipulate the data flow of the system~\cite{Galbally2010PR}. Physical attacks, also known as spoofing attacks, refer to face manipulations made before the data acquisition~\cite{spoofing2014fierrez,galbally14TIP}. The most common physical attack is the usage of tools (e.g. masks or makeup) to change our appearance \cite{hadid15SPMspoofing,2019_BookPAD2_IntroFacePAD_JHO}. As shown in Fig.~\ref{fig:changes_app}, the usage of accessories to disguise your real identity (e.g. scarf or glasses) directly affects the output of a face recognition system, a fact largely studied in forensic face recognition~\cite{2015_FSI_FacialSoftBio_Pedro}. However, we can also observe another easy way to interfere the output of the Face Recognition system, whose impact is at least comparable to the usage of disguising tools: changing your facial expression. Although the influence of facial expressions on face recognition performance is well-known~\cite{chang2005evaluation}, they have typically been studied as another covariate associated to pose variations~\cite{qian2019unsupervised}. To the best of our knowledge, there are no works analyzing the impact of facial expression bias in current deep face recognition technology, and the vulnerability they may suppose to such systems.

In this work, we focus on the prototypical facial expressions of the $6$ basic emotions: happiness, sadness, anger, surprise, disgust, and fear. In Sec.~\ref{sec:material} we describe the different databases and algorithms used in our experiments. Then, we quantify the significant facial expression biases present in the most popular face recognition datasets (Sec. \ref{sec:bias_dataset}). To understand to what extent these facial expression biases affect the performance of face recognition models we performed a series of experiments in controlled scenarios. Our results (Sec.~\ref{sec:experiments}) demonstrate that popular face models trained solely for face identity recognition have self-learned emotional responses\footnote{An emotional response is defined here as an output of a pre-trained model that varies depending of the face expression of the input image~\cite{Ekman2002FACS}.}. We evaluate empirically the impact of such self-learned emotional responses in the security of face recognition systems using $3$ public databases and $3$ face recognition models.


\section{Data and Methods}\label{sec:material}

In order to conduct a comprehensive study of the impact of facial expressions on face recognition systems, we carry out our experiments using different state-of-the-art face recognition pre-trained models: 

\begin{itemize}
    \setlength\itemsep{0pt}
    \setlength{\parskip}{0pt}
    \item VGG$16$~\cite{VGG}: This model has $5$ convolutional blocks that use small convolutional filters in all layers to effectively increase the depth of the network. The model was trained with the VGGFace$2$ database~\cite{VGG2} and the MTCNN face detector~\cite{MTCNN}. 
    \item ResNet-$50$~\cite{ResNet}: This CNN comprises $50$ layers including residual connections. We use the model trained for face recognition with VGGFace$2$~\cite{VGG2} and the MTCNN face detector~\cite{MTCNN}.
    \item LResNet$100$E-IR~\cite{ArcFace}: This model is a residual network~\cite{ResNet} with $44$M parameters, which use a more advanced residual unit setting~\cite{ResNet_Pyramid}. The model was trained with ArcFace loss, the MS-Celeb-$1$M ArcFace dataset~\cite{ArcFace} and the RetinaFace detector~\cite{RetinaFace}.
    
\end{itemize}

All three models were used as feature extractors, computing the matching score between two face images as the Euclidean distance between its two feature vectors. We used three public databases to evaluate the effect of changes in facial expression in face recognition performance: CFEE~\cite{CFEE}, CK+~\cite{CK+}, and CelebA~\cite{CelebA}. Note that, while CelebA is a large database collected using search engines, both CFEE and CK+ were collected under lab-controlled conditions, and therefore they are ideal to analyze the isolated effect of facial expression on the face recognition pipeline.



\begin{table*}[t!]
    \centering
    \small
    \begin{tabular}{l|c|c|c|c|c|c|c|c}
    \hline
    \textbf{Database} & \textbf{\#images} & \textbf{Neutral} & \textbf{Happy} & \textbf{Sad} & \textbf{Anger} & \textbf{Surprised} & \textbf{Disgusted} & \textbf{Fear}\\
    \hline
    MS-Celeb-$1$M \cite{MS-Celeb} & $8.5$M & $83.7\%$ & $5.7\%$ & $0.2\%$ & $3.4\%$ & $2.2\%$ & $4.6\%$ & $\sim0.0\%$\\
    
    MegaFace \cite{2016Megaface} & $4.7$M  & $82.0\%$ & $4.5\%$ & $0.1\%$ & $7.0\%$ & $1.3\%$ & $5.0\%$ & $\sim0.0\%$\\
    
    
    
    CelebA \cite{CelebA} & $203$K  & $62.2\%$ & $33.3\%$ & $0.1\%$ & $0.5\%$ & $1.6\%$ & $0.9\%$ & $\sim0.0\%$\\
    
    IJB-C \cite{IJB-C}       & $21$K & $66.2\%$ & $26.9\%$ & $0.1\%$ & $0.6\%$ & $2.6\%$ & $2.0\%$ & $0.1\%$\\
    
    
    LFW \cite{LFW}         & $13$K  & $61.2\%$ & $28.0\%$ & $0.3\%$ & $1.8\%$ & $3.1\%$ & $4.4\%$ & $0.0\%$\\
    
    
    VGGFace$2$ \cite{VGG2}    & $3.3$M  & $64.5\%$ & $28.2\%$ & $0.2\%$ & $0.4\%$ & $3.3\%$ & $2.0\%$ & $0.1\%$\\
    \hline
    \end{tabular}
    \caption{Facial expression labels of images in the most popular state-of-the-art face databases. The facial expressions were labeled using the Affectiva software (https://www.affectiva.com/).}
    \label{tab:SOA_Databases}
\end{table*}

\section{Expression Bias in Face Datasets}\label{sec:bias_dataset}

One of the main challenges for training a face recognition system based on Deep Models is the need for millions of images from thousands of users involving realistic acquisition conditions (e.g. changes in pose, illumination, or image resolution). This led to the design of large-scale databases using search engines and semi-automated processes to collect images from the Internet, movies or social media in an unconstrained way (e.g. VGGFace2~\cite{VGG2} or MS-Celeb-1M~\cite{MS-Celeb}). Although these databases have significantly contributed to make great advances in the field, 
they may present biases that can impact the performance of the trained systems.


Inspired by the work of~\cite{serna2020sensitiveloss}, where the authors analyze the demographic statistics of multiple face databases, we conducted a similar experiment to study the facial expression bias in popular face databases. For that, we used the COST Affectiva, whose reliability in the expression classification task has already been tested~\cite{kulke2020comparison}, to classify database images into $7$ facial expressions ($6$ basic emotions plus neutral face). Our hypothesis is that the source of the images of these databases can be biased towards certain facial expressions, with some expressions significantly more frequent than others. Tab.~\ref{tab:SOA_Databases} summarizes the facial expression distribution in each database. Note that there is a strong facial expression bias in all the evaluated databases. In all cases, the proportion of Neutral images exceeds $60\%$, reaching a maximum value of $83.7\%$ in the MS-Celeb-$1$M database~\cite{MS-Celeb}. The second most common facial expression is Happy. In fact, for all the datasets, $\sim90\%$ of the images show either Neutral or Happy. 


As for the other $5$ facial expressions, notice that Surprised and Disgusted rarely exceed $6\%$, while Sad, Fear, and Anger have in general a very low representation, often below $1\%$.  
This remarkable under-representation of some facial expressions produces the following drawbacks: i) on the one hand, models are trained using highly biased data that result in heterogeneous performances; ii) on the other hand, technology is evaluated only for mainstream expressions hiding its real performance for images with some specific facial expressions. 



\begin{figure*}[t]
    \centering
    \begin{minipage}{0.33\textwidth}
    \includegraphics[width=\linewidth]{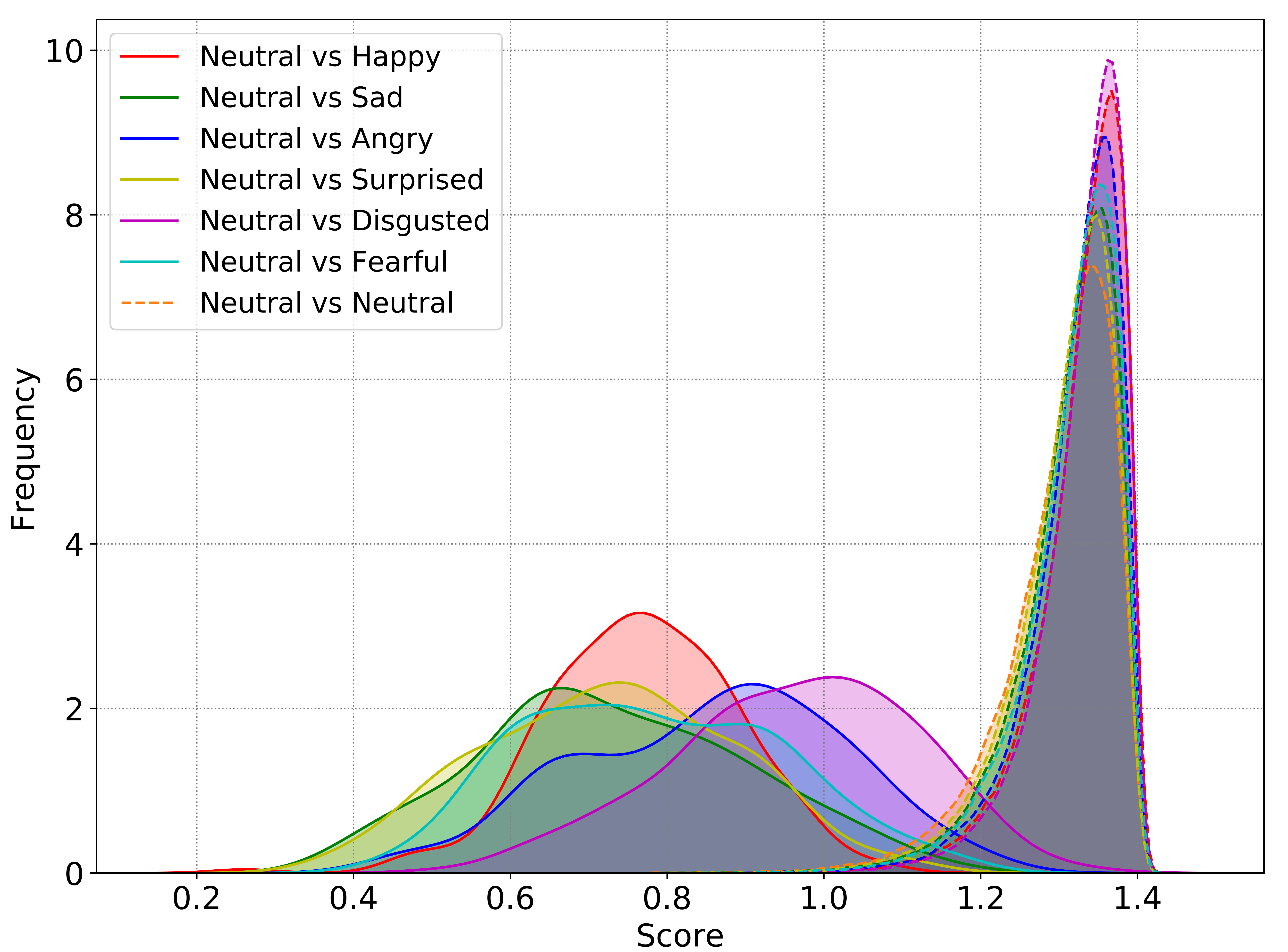}
    \end{minipage}
    \begin{minipage}{0.33\textwidth}
    \includegraphics[width=\linewidth]{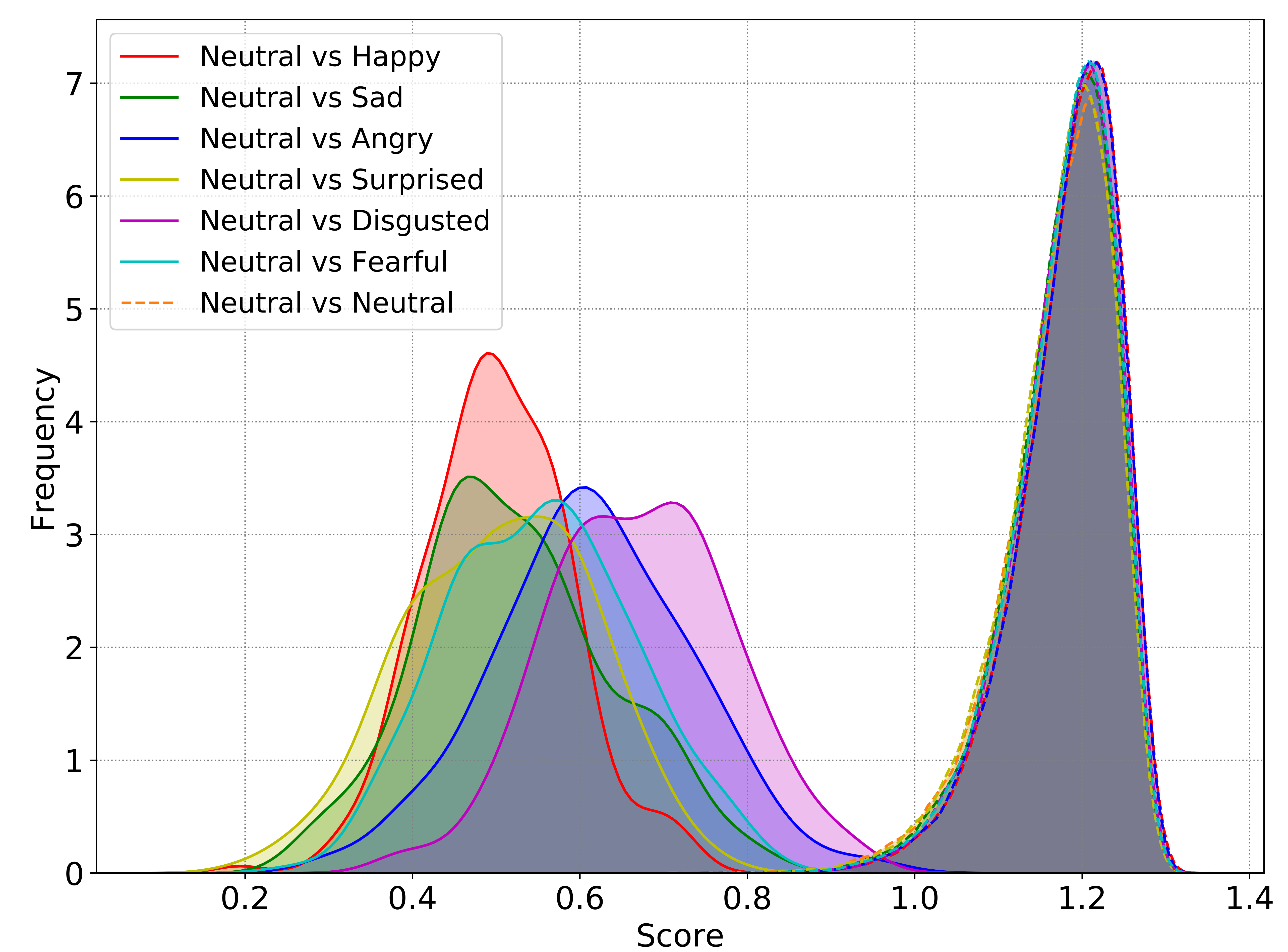}
    \end{minipage}
    \begin{minipage} {0.33\textwidth}
    \includegraphics[width=\linewidth]{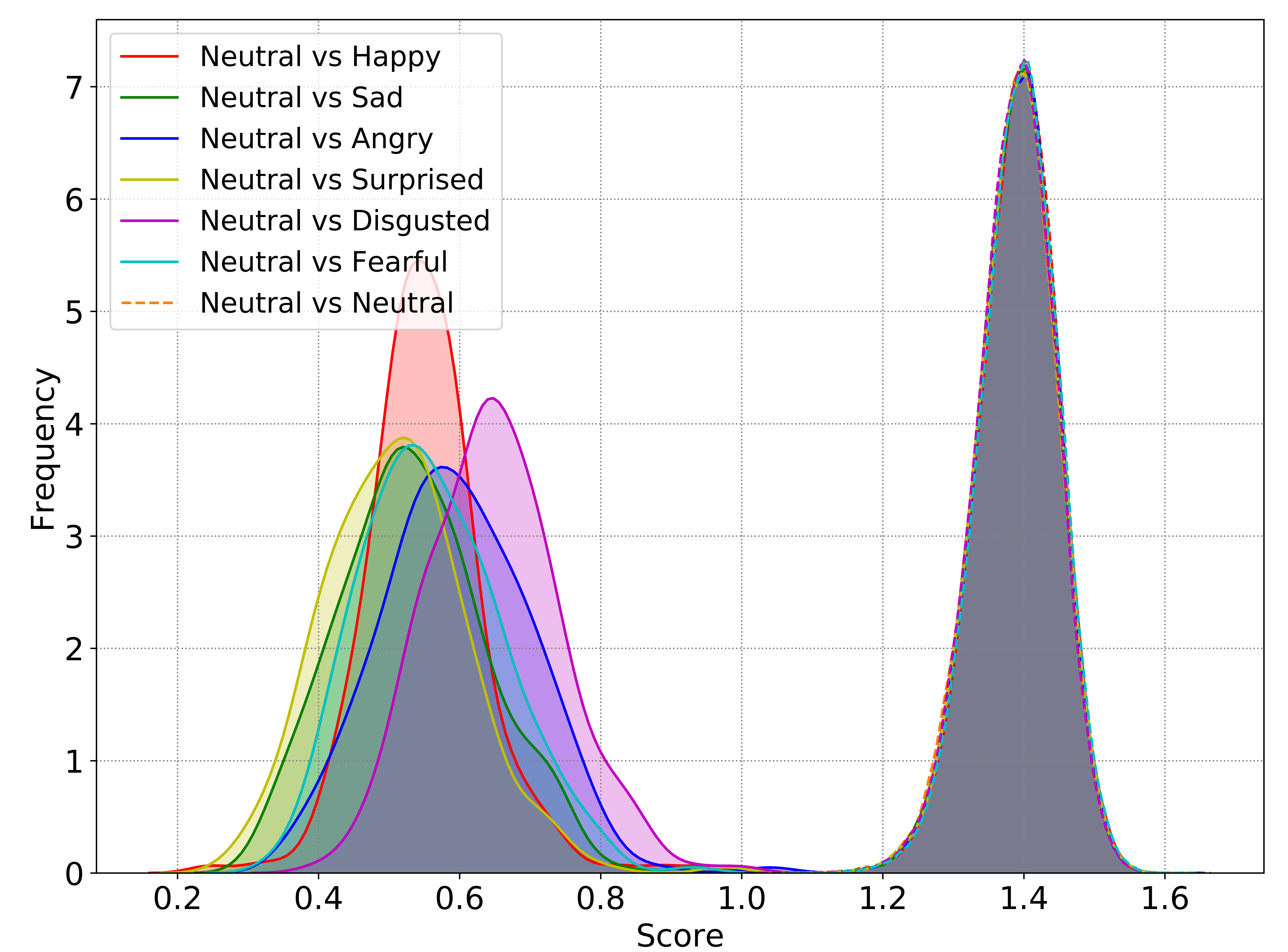}
    \end{minipage}
    \vspace{0pt}
    \caption{Genuine (continuous line) and impostor (dashed line) matching score distributions by facial expression on the CFEE database~\cite{CFEE} for the face matchers: (Left) VGG$16$~\cite{VGG}, (Center) ResNet-$50$~\cite{ResNet}, and (Right) LResNet$100$E-IR~\cite{ArcFace}. }
    \label{fig:dis_neutral}
\end{figure*}

\section{Impact of Facial Expression in Face Recognition Systems}\label{sec:experiments}


Our hypothesis is that the facial expression biases of the training set can impact the robustness of the face recognition systems. Concretely, the biases revealed in  Sec.~\ref{sec:bias_dataset} make us think that the models will work better for faces showing Neutral or Happy expressions, which are the most common expressions. 

We start our analysis on CFEE~\cite{CFEE}, which contain images collected in a very controlled environment with uniform illumination, controlled pose, and without occlusions. Each user in CFEE presents one image unequivocally representing each of the $22$ expressions in the dataset. Thus, the  impact of the facial expression can be isolated from the rest of covariates (e.g. pose, demographic attributes, illumination). We conducted an authentication experiment including seven facial expressions (i.e. Neutral expression and the six basic expressions defined in the FACS manual~\cite{Ekman2002FACS}), where we analyze both genuine and impostor scores distributions by comparing each facial expression with the rest. CFEE is composed of $230$ users, all of them having an image representing each expression. Thus, all the distributions were computed using the same number of image pairs: $230$ for each genuine distribution and $230\times229$ for each impostor distribution.
\begin{table}[t]
    \centering
    \footnotesize
    \begin{tabular}{l|c|c|c|c|c|c}
    \hline
    \multirow{2}{*}{\textbf{Met.}}&\multicolumn{6}{c}{\textbf{Av. Genuine Score \textsubscript{Rank-1 in \%} (Ref. Expression = N) }}\\
    \cline{2-7}
    &\textbf{H}&\textbf{S}&\textbf{A}&\textbf{Su}&\textbf{D}&\textbf{F}\\
    \hline
    (a) &$.77_{96.0}$&$.73_{90.3}$&$.86_{77.3}$&$.73_{89.4}$&$.96_{50.3}$&$.78_{84.2}$\\
    
    (b) &$.50_{100}$&$.52_{100}$&$.61_{97.1}$&$.50_{100}$&$.67_{94.4}$&$.55_{98.8}$\\
    
    (c) &$.55_{100}$&$.53_{100}$&$.59_{100}$&$.51_{99.7}$&$.65_{99.3}$&$.56_{100}$\\
    
    \hline
    \multirow{2}{*}{\textbf{Met.}}&\multicolumn{6}{c}{\textbf{Av. Genuine Score \textsubscript{Rank-1 in \%} (Ref. Expression = H) }}\\
    \cline{2-7}
    &\textbf{N}&\textbf{S}&\textbf{A}&\textbf{Su}&\textbf{D}&\textbf{F}\\
    \hline
    (a)&$.77_{96.9}$&$.87_{84.3}$&$.95_{60.0}$&$.88_{76.4}$&$.97_{55.2}$&$.86_{84.8}$\\
    
    (b)&$.50_{100}$&$.62_{97.6}$&$.67_{97.3}$&$.62_{97.6}$&$.68_{93.0}$&$.60_{98.4}$\\
    
    (c)&$.55_{100}$&$.65_{99.6}$&$.68_{99.6}$&$.64_{99.6}$&$.69_{99.3}$&$.63_{100}$\\
    
    \hline
    \multirow{2}{*}{\textbf{Met.}}&\multicolumn{6}{c}{\textbf{Av. Genuine Score \textsubscript{Rank-1 in \%} (Ref. Expression = S) }}\\
    \cline{2-7}
    &\textbf{H}&\textbf{N}&\textbf{A}&\textbf{Su}&\textbf{D}&\textbf{F}\\
    \hline
    (a)&$.87_{73.4}$&$.73_{94.2}$&$.77_{86.8}$&$.84_{79.1}$&$.92_{56.4}$&$.74_{92.1}$\\
    
    (b)&$.62_{97.2}$&$.52_{99.5}$&$.56_{98.0}$&$.61_{97.6}$&$.66_{89.2}$&$.54_{99.3}$\\
    
    (c)&$.65_{99.2}$&$.54_{100}$&$.55_{99.6}$&$.60_{99.6}$&$.65_{99.2}$&$.56_{100}$\\
    
    \hline
    \multirow{2}{*}{\textbf{Met.}}&\multicolumn{6}{c}{\textbf{Av. Genuine Score \textsubscript{Rank-1 in \%} (Ref. Expression = A) }}\\
    \cline{2-7}
    &\textbf{H}&\textbf{S}&\textbf{N}&\textbf{Su}&\textbf{D}&\textbf{F.}\\
    \hline
    (a)&$.95_{54.1}$&$.77_{88.0}$&$.86_{80.4}$&$.96_{44.4}$&$.87_{69.1}$&$.90_{78.4}$\\
    
    (b)&$.67_{95.3}$&$.56_{97.1}$&$.61_{97.4}$&$.71_{86.6}$&$.62_{83.5}$&$.65_{95.6}$\\
    
    (c)&$.68_{100}$&$.55_{100}$&$.59_{100}$&$.66_{99.7}$&$.60_{99.2}$&$.63_{100}$\\
    
    \hline
    \multirow{2}{*}{\textbf{Met.}}&\multicolumn{6}{c}{\textbf{Av. Genuine Score \textsubscript{Rank-1 in \%} (Ref. Expression = Su) }}\\
    \cline{2-7}
    &\textbf{H}&\textbf{S}&\textbf{A}&\textbf{N}&\textbf{D}&\textbf{F}\\
    \hline
    (a)&$.88_{62.1}$&$.84_{74.7}$&$.96_{43.7}$&$.72_{81.0}$&$.99_{37.3}$&$.73_{66.0}$\\
    
    (b)&$.62_{95.1}$&$.61_{95.6}$&$.71_{87.0}$&$.50_{96.5}$&$.71_{83.5}$&$.52_{90.5}$\\
    
    (c)&$.64_{99.7}$&$.60_{100}$&$.66_{99.6}$&$.51_{99.9}$&$.70_{99.3}$&$.53_{100}$\\
    
    \hline
    \multirow{2}{*}{\textbf{Met.}}&\multicolumn{6}{c}{\textbf{Av. Genuine Score \textsubscript{Rank-1 in \%} (Ref. Expression = D) }}\\
    \cline{2-7}
    &\textbf{H}&\textbf{S}&\textbf{A}&\textbf{Su}&\textbf{N}&\textbf{F}\\
    \hline
    (a)&$.97_{41.9}$&$.92_{53.9}$&$.87_{65.5}$&$.99_{35.8}$&$.96_{40.4}$&$.95_{43.0}$\\
    
    (b)&$.68_{86.0}$&$.66_{85.0}$&$.62_{93.5}$&$.71_{77.8}$&$.67_{90.3}$&$.67_{83.4}$\\
    
    (c)&$.69_{99.3}$&$.65_{99.6}$&$.60_{99.2}$&$.70_{99.3}$&$.65_{99.7}$&$.67_{100}$\\
    
    \hline
    \multirow{2}{*}{\textbf{Met.}}&\multicolumn{6}{c}{\textbf{Av. Genuine Score \textsubscript{Rank-1 in \%} (Ref. Expression = F) }}\\
    \cline{2-7}
    &\textbf{H}&\textbf{S}&\textbf{A}&\textbf{Su}&\textbf{D}&\textbf{N}\\
    \hline
    (a)&$.86_{75.2}$&$.74_{91.1}$&$.90_{61.9}$&$.73_{88.0}$&$.95_{52.9}$&$.78_{83.0}$\\
    
    (b)&$.60_{98.8}$&$.54_{99.2}$&$.65_{92.5}$&$.52_{98.4}$&$.67_{87.6}$&$.55_{98.6}$\\
    
    (c)&$.63_{99.2}$&$.56_{99.6}$&$.63_{99.2}$&$.53_{99.6}$&$.67_{99.2}$&$.56_{99.4}$\\
    \hline
    \end{tabular}
    
    \caption{Average genuine score in CFEE~\cite{CFEE}, and Rank-$1$ statistics (\% in subscript) in CFEE, CK+~\cite{CK+} and CelebA~\cite{CelebA} with different face expressions as reference for the face matchers: (a) VGG$16$~\cite{VGG}, (b) ResNet-50~\cite{ResNet}, and (c) LResNet100E-IR~\cite{ArcFace}.}
    \label{tab:rank1}
\end{table}

Fig.~\ref{fig:dis_neutral} presents the genuine and impostor matching scores distributions (pairs of faces), using Neutral expression as reference, computed with the three face recognition algorithms presented in Sec.~\ref{sec:material}. Notice that, while the genuine distributions do not include the case Neutral vs Neutral due to the limitations of CFEE (only one image per expression), the impostor distributions contemplate this scenario. We observe that, while the genuine distributions are clearly influenced by the facial expressions being compared, the impostor distributions barely change across expressions. In both ResNet-$50$ and LResNet$100$E-IR the impostor distributions are practically identical. The impostor distributions obtained with VGG$16$ show more overlap than the others, probably because the poorest performance of that model in comparison with ResNet-$50$ and LResNet$100$E-IR. At the same time, if we focus on the genuine distributions in Fig.~\ref{fig:dis_neutral}, we clearly observe a difference among expressions in all three cases. For the rest of experiments we will focus on these genuine differences.

Tab.~\ref{tab:rank1} presents the average genuine score for each distribution. Considering Neutral as the reference expression, the Happy, Sad, and Surprised distributions show the lowest scores, while the Disgusted and Anger distributions tend to higher values. For the three models in Fig.~\ref{fig:dis_neutral} we observe that all the distributions seem to have a similar deviation, except for the case of Happy, which shows a higher concentration of scores around the average. LResNet$100$E-IR is a state-of-the-art face recognition model with very competitive performances in a large list of databases and protocols. However, the results demonstrate that its output is largely affected by the facial expression, with genuine score variations of more than $30\%$ across expressions in some cases. On the other hand, several distributions computed by VGG$16$ exceed the score threshold of $1$, where the impostors begin, having a large overlap in the Anger and Disgust distributions. 


Regarding the genuine score when non-neutral facial expressions are used as the reference (see Tab.~\ref{tab:rank1}), the observed trend is similar to the tend observed with the Neutral expression. The genuine distributions depend on the expressions compared in all three methods, with VGG$16$ having the greatest vulnerability to this effect (i.e. higher average score). In general, Anger, Sad, and Fear expressions work well together. Surprise expression has a better affinity with Fear and Neutral expressions, while Disgusted ones obtain the worst results with almost all expressions. Happy expressions seem to work well with Neutral expressions (i.e. the two most common expressions in training databases, as analyzed in Sec.~\ref{sec:bias_dataset}).


This impact in the genuine matching score distributions suppose a potential vulnerability, as it may influence the probabilities to be correctly identified by these systems. To study this possibility, we designed an identification experiment with the subjects in CFEE and CK+~\cite{CK+}, a total of $350$ different identities and $16$K different images. In order to increase the background set, we also add a set of $15$K images from the CelebA~\cite{CelebA}. We are aware that the characteristics of CelebA dataset are very different to those from CFEE or CK+. However, CelebA includes good quality images that can serve to increase the difficulty of the experiment. For each user in CFEE, we selected a specific expression image as query sample, computing the matching score with all the images belonging to the rest of users in the background set. We then compare the background scores with each of the user's genuine scores, computed by comparing the query sample with the other $6$ basic facial expressions, and extract Rank-$1$ statistics individually for each facial expression.


Tab.~\ref{tab:rank1} presents Rank-$1$ results by facial expression, using the three face recognition algorithms with their respective face detectors. We observe lower Rank-$1$ results when using VGG$16$ in all cases. Both ResNet-$50$ and LResNet$100$E-IR obtained good results, being slightly higher in the second one. The Rank-$1$ results follow the trend seen in the case of the average genuine scores, maintaining the relationships between expressions previously mentioned. The Disgusted expressions obtained the worst results in all cases, confirming the low affinity with the other expressions that we observed in the scores distributions. The drop of performance is clear for VGG-$16$ and ResNet-$50$ models, with Rank-1 differences between expressions up to $50\%$. These differences are moderate for the LResNet$100$E-IR model. The high quality standards of CFEE images are not a challenge to a state-of-the-art model. However, the drop of performance in LResNet$100$E-IR exists and varies between $0-1\%$, even in this simple environment. It is expected that this impact increases in a more challenging benchmark where expressions and other covariates affect the performance at the same time.

\section{Conclusions}\label{sec:conclusions}
This work analyzes the impact of facial expressions bias on face recognition systems. 
Our main findings are: i) The most popular face recognition databases systematically present huge facial expression biases. Therefore, new public databases are needed to overcome the scarce representation of some facial expressions in existing databases. ii) Facial expression bias affects the performance of genuine comparisons with variations in scores of up to $40\%$. Facial expression bias does not affect impostor comparisons. Thus, we can discard impersonation attacks based on facial expression manipulation. However, obfuscation of identity based on manipulation the face expression is easy and particularly harmful.

As a result of this work, we strongly advocate for reducing the facial expression bias in future face recognition databases, and further development of bias-reduction methods applicable to existing databases and existing models already trained on biased datasets~\cite{2020_DeBias_Jain,2021_TPAMI_SensitiveNets_Morales,ter2021comprehensive}. The application of face manipulation techniques~\cite{nirkin2019fsgan,tolosana2020deepfakes,2020_JSTSP_GANprintR_Neves} could serve to enhance the facial expression availability on existing database by introducing new synthetically generated expressions. Analyzing varying facial expressions in time is also in our plans \cite{2020_ICIP_QIDmulti_Perera}.


\section{Acknowledgments}

Supported by projects: TRESPASS-ETN (MSCA-ITN-2019-860813), PRIMA (MSCA-ITN-2019-860315), BIBECA (RTI2018-101248-B-I00 MINECO/FEDER), and RTI2018-095232-B-C22. A. Pe\~na is supported by a research fellowship from MINECO.

{\small
\bibliographystyle{IEEEbib}
\bibliography{bibliography}

\begin{thebibliography}{10}

\bibitem{patel18spm_face}
R.~{Ranjan} et~al.,
\newblock ``Deep learning for understanding faces: Machines may be just as
  good, or better, than humans,''
\newblock {\em IEEE Signal Processing Magazine}, vol. 35, no. 1, pp. 66--83,
  2018.

\bibitem{2018_TIFS_SoftWildAnno_Sosa}
E.~Gonzalez-Sosa, J.~Fierrez, et~al.,
\newblock ``Facial soft biometrics for recognition in the wild: Recent works,
  annotation and cots evaluation,''
\newblock {\em IEEE Trans. on Information Forensics and Security}, vol. 13, no.
  8, pp. 2001--2014, August 2018.

\bibitem{lu2019experimental}
B.~Lu, J.~Chen, C.~D. Castillo, and R.~Chellappa,
\newblock ``An experimental evaluation of covariates effects on unconstrained
  face verification,''
\newblock {\em IEEE Trans. on Biometrics, Behavior, and Identity Science}, vol.
  1, no. 1, pp. 42--55, 2019.

\bibitem{ResNet}
K.~{He}, X.~{Zhang}, S.~{Ren}, and J.~{Sun},
\newblock ``{Deep residual learning for image recognition},''
\newblock in {\em IEEE CVPR}, 2016.

\bibitem{Galbally2007_Vulnerabilities}
J.~Galbally, J.~Fierrez, and J.~Ortega-Garcia,
\newblock ``Vulnerabilities in biometric systems: {A}ttacks and recent advances
  in liveness detection,''
\newblock in {\em Proc. Spanish Workshop on Biometrics}, 2007.

\bibitem{Galbally2010PR}
J.~Galbally, C.~McCool, J.~Fierrez, et~al.,
\newblock ``On the vulnerability of face verification systems to hill-climbing
  attacks,''
\newblock {\em Pattern Recognition}, vol. 43, no. 3, pp. 1027--1038, March
  2010.

\bibitem{spoofing2014fierrez}
J.~Galbally, S.~Marcel, and J.~Fierrez,
\newblock ``Biometric antispoofing methods: A survey in face recognition,''
\newblock {\em IEEE Access}, vol. 2, pp. 1530--1552, 2014.

\bibitem{galbally14TIP}
J.~Galbally, S.~Marcel, and J.~Fierrez,
\newblock ``Image quality assessment for fake biometric detection: Application
  to iris, fingerprint and face recognition,''
\newblock {\em IEEE Trans. on Image Processing}, vol. 23, no. 2, pp. 710--724,
  February 2014.

\bibitem{hadid15SPMspoofing}
A.~Hadid, N.~Evans, et~al.,
\newblock ``{Biometrics systems under spoofing attack: An evaluation
  methodology and lessons learned},''
\newblock {\em IEEE Signal Processing Magazine}, vol. 32, no. 5, pp. 20--30,
  2015.

\bibitem{2019_BookPAD2_IntroFacePAD_JHO}
J.~Hernandez-Ortega, J.~Fierrez, A.~Morales, and J.~Galbally,
\newblock ``Introduction to face presentation attack detection,''
\newblock in {\em Handbook of Biometric Anti-Spoofing}, S.~Marcel et~al., Eds.,
  pp. 187--206. Springer, 2019.

\bibitem{2015_FSI_FacialSoftBio_Pedro}
P.~Tome, R.~Vera-Rodriguez, J.~Fierrez, and J.~Ortega-Garcia,
\newblock ``Facial soft biometric features for forensic face recognition,''
\newblock {\em Forensic Science International}, vol. 257, pp. 171--284, 2015.

\bibitem{chang2005evaluation}
K.~Chang, K.~W. Bowyer, and P.~J. Flynn,
\newblock ``An evaluation of multimodal 2{D} + 3{D} face biometrics,''
\newblock {\em IEEE Trans. on Pattern Analysis and Machine Intelligence}, vol.
  27, pp. 619--624, 2005.

\bibitem{qian2019unsupervised}
Y.~Qian, W.~Deng, and J.~Hu,
\newblock ``Unsupervised face normalization with extreme pose and expression in
  the wild,''
\newblock in {\em IEEE CVPR}, 2019.

\bibitem{Ekman2002FACS}
P.~Ekman, W.~Friesen, and J.~Hager,
\newblock {\em {The Facial Action Coding System}},
\newblock Salt Lake City, UT, 2002.

\bibitem{VGG}
K.~Simonyan and A.~Zisserman,
\newblock ``Very deep convolutional networks for large-scale image
  recognition,''
\newblock in {\em ICLR}, 2015.

\bibitem{VGG2}
Q.~Cao, L.~Shen, et~al.,
\newblock ``{VGGFace2: A dataset for recognising faces across pose and age},''
\newblock in {\em IEEE FG}, 2018.

\bibitem{MTCNN}
K.~{Zhang}, Z.~{Zhang}, et~al.,
\newblock ``Joint face detection and alignment using multitask cascaded
  convolutional networks,''
\newblock {\em IEEE Signal Processing Letters}, vol. 23, no. 10, pp.
  1499--1503, 2016.

\bibitem{ArcFace}
J.~{Deng}, J.~{Guo}, et~al.,
\newblock ``{ArcFace: Additive angular margin loss for deep face
  recognition},''
\newblock in {\em IEEE CVPR}, 2019.

\bibitem{ResNet_Pyramid}
D.~{Han}, J.~{Kim}, and J.~{Kim},
\newblock ``Deep pyramidal residual networks,''
\newblock in {\em IEEE CVPR}, 2017.

\bibitem{RetinaFace}
J.~Deng, J.~Guo, Ververas, et~al.,
\newblock ``{RetinaFace: Single-shot multi-level face localisation in the
  wild},''
\newblock in {\em IEEE CVPR}, 2020.

\bibitem{CFEE}
S.~Du, Y.~Tao, and A.~M. Martinez,
\newblock ``Compound facial expressions of emotion,''
\newblock {\em Proc. of the National Academy of Sciences}, vol. 111, pp.
  1454--1462, 2014.

\bibitem{CK+}
P.~{Lucey}, J.~F. {Cohn}, et~al.,
\newblock ``{The extended Cohn-Kanade dataset (CK+): A complete dataset for
  action unit and emotion-specified expression},''
\newblock in {\em IEEE CVPRW}, 2010.

\bibitem{CelebA}
Z.~Liu, P.~Luo, X.~Wang, and X.~Tang,
\newblock ``Deep learning face attributes in the wild,''
\newblock in {\em ICCV}, 2015.

\bibitem{MS-Celeb}
Y.~Guo, L.~Zhang, Y.~Hu, X.~He, and J.~Gao,
\newblock ``{MS-Celeb-1M: A dataset and benchmark for large-scale face
  recognition},''
\newblock in {\em IEEE ECCV}, 2016.

\bibitem{2016Megaface}
I.~Kemelmacher-Shlizerman, S.~M. Seitz, D.~Miller, and E.~Brossard,
\newblock ``{The MegaFace benchmark: 1 million faces for recognition at
  scale},''
\newblock in {\em IEEE CVPR}, 2016.

\bibitem{IJB-C}
B.~{Maze}, J.~{Adams}, et~al.,
\newblock ``{IARPA Janus Benchmark - C: Face Dataset and Protocol},''
\newblock in {\em IAPR ICB}, 2018.

\bibitem{LFW}
G.~B. Huang, M.~Ramesh, T.~Berg, and E.~Learned-Miller,
\newblock ``Labeled faces in the wild: A database for studying face recognition
  in unconstrained environments,''
\newblock Tech. {R}ep. 07-49, University of Massachusetts, Amherst, October
  2007.

\bibitem{serna2020sensitiveloss}
I.~Serna, A.~Morales, J.~Fierrez, M.~Cebrian, N.~Obradovich, and I.~Rahwan,
\newblock ``{SensitiveLoss: Improving accuracy and fairness of face
  representations with discrimination-aware deep learning},''
\newblock {\em arXiv/2004.11246}, 2020.

\bibitem{kulke2020comparison}
L.~Kulke et~al.,
\newblock ``{A comparison of the Affectiva iMotions facial expression analysis
  software with EMG for identifying facial expressions of emotion},''
\newblock {\em Frontiers in Psychology}, 2020.

\bibitem{2020_DeBias_Jain}
S.~Gong, X~Liu, and A.~K. Jain,
\newblock ``Jointly de-biasing face recognition and demographic attribute
  estimation,''
\newblock in {\em IEEE ECCV}, 2020, pp. 330--347.

\bibitem{2021_TPAMI_SensitiveNets_Morales}
A.~Morales, J.~Fierrez, R.~Vera-Rodriguez, and R.~Tolosana,
\newblock ``{SensitiveNets}: Learning agnostic representations with application
  to face recognition,''
\newblock {\em IEEE Trans. on Pattern Analysis and Machine Intelligence}, vol.
  43, pp. 2158--2164, 2021.

\bibitem{ter2021comprehensive}
P.~Terhorst et~al.,
\newblock ``A comprehensive study on face recognition biases beyond
  demographics,''
\newblock {\em arXiv/2103.01592}, 2021.

\bibitem{nirkin2019fsgan}
Y.~Nirkin, Y.~Keller, and T.~Hassner,
\newblock ``{FSGAN}: Subject agnostic face swapping and reenactment,''
\newblock in {\em ICCV}, 2019.

\bibitem{tolosana2020deepfakes}
R.~Tolosana, R.~Vera-Rodriguez, J.~Fierrez, et~al.,
\newblock ``{DeepFakes} and beyond: A survey of face manipulation and fake
  detection,''
\newblock {\em Information Fusion}, vol. 64, pp. 131--148, 2020.

\bibitem{2020_JSTSP_GANprintR_Neves}
J.C. Neves et~al.,
\newblock ``{GANprintR}: Improved fakes and evaluation of the state of the art
  in face manipulation detection,''
\newblock {\em IEEE Journal of Selected Topics in Signal Processing}, vol. 14,
  no. 5, pp. 1038--1048, August 2020.

\bibitem{2020_ICIP_QIDmulti_Perera}
P.~Perera, J.~Fierrez, and V.~Patel,
\newblock ``Quickest intruder detection for multiple user active
  authentication,''
\newblock in {\em IEEE Intl. Conf. on Image Processing (ICIP)}, October 2020.

\end{thebibliography}
}

\end{document}